# Subjective Reality and Strong Artificial Intelligence

Alexander Serov

Machinery Research Institute, Russian Academy of Sciences, Moscow, Russian Federation

E-mail: alexser1929@gmail.com

## Abstract

The main prospective aim of modern research related to Artificial Intelligence is the creation of technical systems that implement the idea of Strong Intelligence. According our point of view the path to the development of such systems comes through the research in the field related to perceptions. Here we formulate the model of the perception of external world which may be used for the description of perceptual activity of intelligent beings. We consider a number of issues related to the development of the set of patterns which will be used by the intelligent system when interacting with environment. The key idea of the presented perception model is the idea of subjective reality. The principle of the relativity of perceived world is formulated. It is shown that this principle is the immediate consequence of the idea of subjective reality. In this paper we show how the methodology of subjective reality may be used for the creation of different types of Strong AI systems.

## Introduction

One of the fastest growing areas of science now is the area of Artificial Intelligence (AI). Apparently it will not be wrong to say that the leadership in the implementation of technologies based on Artificial Intelligence belongs to DARPA [13, 16, 20, 3]. Emerging civilian [5] and public [1, 32] types of AI applications make it easy to forecast the growing interest to research in the field of automated analytical data processing in near future. This interest is actively stimulated by the rapid growth of information available to any member of human society due to the development of technologies of global networks [2].

The application of modern numerical methods that are traditionally related to the field of artificial intelligence today is very diverse (see, for example [24]). However up to now the issue of the intelligence remains unclear. Ordinary [24] the definition of the intelligence is associated with the ability to thought processes, reasoning and rational behavior. The concept of intelligence is often associated with the human form of life through the use of the Turing test. According our point of view this unfairly restricts the scope of application of the hypothesis on appropriate and purposeful behavior of living beings which is based on the preliminary experience and learning.

The main prospective aim of modern research related to Artificial Intelligence is the creation of technical systems that implement the idea of Strong Intelligence. At present the research and development in this field is exclusively extensive (see, for example, [31, 8, 9, 6, 28, 7, 33, 11, 4, 10, 27]). Leading position in this direction of work currently belongs to the direction associated with Computer Vision and Visual Intelligence. This is not accidental. Now the science of AI is coming through the way of the development of technologies reproducing the means of the interacting with outer world which are inherent to human beings. And most of the information about the world people gets through visual perception. According our point of view the path to a successful solution of the problem of systems creation that have intelligent abilities similar to human ones comes through the research in the field related to perceptions. Through the way of investigation of problem how living beings are developing their capacity for intellectual activity, through the understanding what is the ideological foundation for the arising of this ability.

The world around us is in constant motion and change. All beings are in constant interaction and change. The interaction of living creatures with the world is based on their unique ability to perceive. Perception, as a form of interaction with the outside world, is one of those abilities inherent for living beings, which generate together the phenomena of the intellect. We believe that this ability is a key element in intellectual activity.

## Perception and Computational Theory of Perception

In modern scientific literature on the subject of numerical methods in the field of Artificial Intelligence there exists the relatively small number of articles on issues of perception. In particular, papers [12, 25, 30, 34, 35, 19, 17, 29] may be mentioned here. These articles are a pioneer in the field which is called Computational Theory of Perception (CTP). Authors of these papers have built the framework for addressing issues of perception in terms of modern computing technology. In order to make the steps in the direction of constructing Strong AI systems it's necessary to deepen the consideration of the issues of perception.

The pioneering papers of Zadeh are dedicated to the representation of information and knowledge. As part of CTP the definition of granules is introduced. Granules are used to describe the results of perception: "The basic unit of meaning is a granule" [30]. "Computational perception is a representation of the information obtained about an object with a level of granularity useful for the programmer's purposes" [30]. From this point of view modern CTP looks more like a science about data representation than the science linking the outer world and the intelligent being perceiving this world. From our point of view the field of science mathematically describing the process of perception of the external world must introduce numerical methods for connecting the process of perception with the data structures that describe the experience and knowledge.

If we describe the intelligence as the ability of some system to a purposeful and expedient activities based on the learning process, we have to conclude the following. Intellectual activity is impossible without the ability to do certain actions in the outside world. Doing activities and watching their results, intelligent system is able to assess and, where necessary to clarify the knowledge of the outside world, which it has at present time. And in this case the methods which intelligent system use to perceive the outward reality plays a very important role in a whole set of activities of the above mentioned system.

## Model of perception of the external world

Let us consider some technical or alive system that is able to perceive the world, make an analysis of the results of perception, consolidate the results of this analysis as a knowledge and use the knowledge to perform some goal-directed activity. For brevity, we call it AI-system. Since the moment when the perception of AI-system starts working, the process of learning starts. Here it is appropriate to mention the model of the three worlds, introduced by Karl Popper [21], [22]. Popper introduced the model of three interacting worlds:

- World of physical objects and events, including, in particular, biological beings
- World of mental objects and events
- World of objective contents of thought

At the time moment when the senses of the outside world of AI-system begin to function, the world of physical objects and phenomena begins to exist for this intelligent system. However, at the same moment of time the world of mental objects and events of this AI-system is empty. This may be considered as one of the tenets of the model of perception. At the time moment of the opening of perception channels AI-system undergoes a kind of information blow. Sensors start sending the stream of data into the system, and this data has not any mapping in the personal world of mental objects which AI-system has at this time.

Opening of channels of perception leads to the need for processing of the data streams coming from sensors into AI-system. The very fact that the processing of perception data streams occurs, is associated with a lack of resources which AI-system has and which can be used to navigate in the outside world. With a reasonable degree of generality we can assume that the computing resources that AI-system has, are not enough for the "complete" data flow processing. This postulate is in a full compliance with current views on the processing of data streams [2]. In general, we can assume that the intensity of the data stream, the resource of AI-system memory and the maximum speed of data flow processing have the values such that the following conditions are true. First, AI-system is not able to process the entire stream of data coming from sensors, but only some of it. Second, the system can't store the entire data stream, but only - some results of the processing of this flow.

Situation related to the time of the opening of perception channels can be characterized by the term "cold start" which is actively used in the fields of AI-oriented numerical methods and Data Mining. The emptiness of the world of mental objects and events, directs the activities of AI-system toward the filling this world as quickly as possible. The primary purpose of this activity is the need to learn to navigate the outside world. We assume that one of the major practical goals which AI-system must achieve at the formation of its personal world of mental objects and phenomena is the goal of reducing energy consumption used for the perception of the outside world. This postulate is consistent with the ideas expressed by Norwich [18, 17]: beings no longer actively perceive the world, after the current situation, the surroundings become clear to them.

There is one issue the solution of which one can significantly advance the development of Artificial Intelligence. This issue concerns the formation of a set of patterns by which AI-system fills its personal world of mental objects and events. In the described model of perception, we introduce the postulate of adaptive constructing of patterns. When we are talking not about alive but technical systems, the memory used by AI-system, may be considered as a constant value. This value is defined by the design of the technical system. The intensity of the data flow coming from the set of sensors into AI-system and maximum speed of the processing of this data flow are defined by the design of the system as well.

Based on this we can conclude that the system of patterns of the outward reality, which is developing by the bearer of intelligence, must be such as to satisfy the following requirements. First, AI-system must be capable of storing the set of patterns. The set of patterns of reality should be as detailed as it is allowed by the resources of AI-system and by the methods of memory using by intelligent system. Secondly, the set of patterns of reality must be such that AI-system will be able to use it during the interaction with the outward world to describe the dynamics of this world. We can describe here the analogy with the nature. If the set of patterns is too detailed, we can see that the presence of a sufficiently low speed of processing can lead to extreme slow response to changes in the external world. A low survival rate of the intelligent beings will be the consequence of this in the presence of aggressive external world. On the other hand, if the set of patterns of the external world is overly simplistic, it will inevitably lead to errors in predicting the dynamics of the observed world. The consequence of this will be similar to the described above. Thus we can conclude that the set of patterns of the external world developed by intelligent systems must maximize their survival.

Application of a priori models, according our opinion, is one of the factors most negatively impacting on the development of Artificial Intelligence. These models are used as a basic set. The whole logic of data processing by intellectual technical system is based on this set of models. We can give here the following example describing the methods currently used in the field of Computer Vision (see, for example, [15, 14, 26]).

Now in this area there are the following two basic approaches to image processing: Appearance-based Methods and Feature-based Methods. First group of methods historically occurred before the second. It is based on the processing of the pixel structure of the image without any a priori information about the observed reality. The result of the use of this technology of image processing usually is characterized by a quite low detectability of persons and other objects of the observable world. As a result, there emerged an image processing area, associated with the Feature-based Methods. These methods use information about the nature of the observed reality, and its features. In particular, methods of the recognition of faces usually use a number of models on the basis of which it is possible to find the differences in the shape and relative positions of the eyes, nose, lips and eyebrows. On the one hand, the result of the application of these methods is characterized by sufficiently high degree of recognition. But on the other hand, the use of these methods is strictly limited by application area: they can be used only for the recognition of human faces. The data used by Feature-based Methods as a priori information must be the result of prior learning of the intellectual technical system.

Using of a priori information about the world, observed by AI-system, not only limits the applicability of the developed numerical methods, but also makes it absolutely impossible to create a Strong AI-system. We have the following point of view. Successful development of a technical system that meets the requirements of Strong AI is possible only in the case when all provisions about objective reality will be removed from the set of models realizing the solution of the problems of intelligent data processing. Any provision based on the concept of objectivity which is used in the presented model of perception will act as some constraint or a model that has an external origin with respect to AI-system. Carriers of highly intelligent functions are able to perceive the world around them. And as part of this perception they can acquire knowledge as a result of generalization, abstraction and validation of results got from the experience of perception of other members of their species. Any instance of technical system that meets the requirements of Strong AI is a personality. Objective reality is given to personality through his feelings, but the sole reality of this personality is the subjective reality.

## Subjective reality of Strong AI-system and the multiplicity of worlds

Subjective reality, within which the intellectual activity of Strong AI-system is placed, is the result of personal experience obtained by this intelligent system. In a section "Experience as a method" [23] Karl Popper gives the following words regarding empirical experience. "Our problem is to formulate an acceptable definition of the term "empirical science" - is not without difficulties. Difficulties stem partly from the fact that, apparently, there are many theoretical systems with logical structure, very similar to the structure of the theoretical framework, which at any given time is given by scientists as this one they adopted as a system of empirical science. Sometimes this situation is described as follows: there is a huge, probably an infinite number of "logically possible worlds". And a system called "empirical science" according its purpose describes the only one world - the "real world" or "world of our experience" (here we present reverse translation from Russian language publication [23]).

Empirical knowledge possessed by a particular instance of AI-system is based on the processing of the data stream coming from the set of sensors that this system has. Between "real world" and "the world of our experience" there cannot be the equal sign for several different reasons. The set of sensors, with which any AI-system perceives the world, is always limited. It is limited both in the number of sensors, and in a set of physical principles on which the work of each individual sensor is based on. From this it follows directly that no system of sensors cannot give a complete "picture of reality". Always some part of the real world will remain outside perception and empirical experience. The perception of an intelligent system only "cuts" some slice of this reality.

In connection with the above, we have to introduce another postulate in a model of perception: the postulate of relativity of the perceived world. The "picture of the world", which is directly perceived by intelligent system, always is determined by the system of sensors by which AI-system is equipped. Relativity of the personal representation of reality has several implications that are important from the point of view of philosophy, and from the point of view of practical application in the field of Artificial Intelligence.

Living beings come into their life helpless, in need of care. The main challenge facing the consciousness at this stage of life is to learn how to navigate in the world subjective picture of which is created by organs of perception. Gradually during the process of learning living creature generates a system of patterns, which most well corresponds to the set of intellectual tasks that are solving by consciousness during the life. This set of patterns seems self-consistent and complete. This illusion arises due to the fact that during the processing of empirical experience a living being assimilates all the laws of subjective reality, which is created by the set of sensors. Nevertheless the incompleteness of the set of patterns will manifest itself in the empirical experience of the living being. This is evidenced by some of the research materials of personality psychology. Incompleteness will manifest itself when trying to introduce a strict mathematical formalism describing the empirical experience of the individual. All of the above also applies to the system of patterns, which is used by a separate species of intelligent beings.

One of the possible practical applications of the postulate of relativity is as follows. A set of sensors, which Strong AI-system can be equipped with, in general case, can be arbitrary. It can combine the sensors, principles of action of which are very diverse. Sensors of this set can be located in space or separated by thousands of miles. Each of the different types of the design of sensors system forms a separate, distinct from the other picture of the outside world for intellectual system, which is equipped with this set of sensors. Thus, from a technical point of view, there is the possibility of formation of very specific types of subjective reality. By working in the boundaries of this subjective reality an intelligent technical system will be able to solve the problems that are important from the point of view of the developers constructing these AI-systems.

## Conclusion

The development of science and its technological applications, of course, allows us to extend the system of sensors, based on data from which humanity creates its modern picture of reality. The creation of the quantum theory has made a revolution in the field of physics. Eventually it makes possible to create a completely new measurement methods, which allowed mankind to enter the world of nano-scale objects and phenomena. Formulated principle of relativity of perceived reality allows us to describe the technical progress of mankind as an expanding of the horizons of our experience.

For a several centuries mankind creates a "second nature." Technical tools that we use in our lives form the technology foundation that makes our life more convenient and comfortable. However, the main aim of the creation of technical means of "second nature" is the reducing in the degree of influence of the nature on the living conditions and the very existence of humanity. Progress towards Strong AI-systems is one of the steps to improve the technological foundation of human society. Creating Strong AI-systems can significantly increase the stability of our living conditions. Perhaps it is at this stage of the development of human society we finally meet with "brothers in mind." Though they will be created artificially.